\documentclass{article}

\usepackage{arxiv}

\usepackage[utf8]{inputenc} 
\usepackage[T1]{fontenc}    
\usepackage{hyperref}       
\usepackage{url}            
\usepackage{booktabs}       
\usepackage{amsfonts}       
\usepackage{nicefrac}       
\usepackage{microtype}      
\usepackage{cleveref}       
\usepackage{lipsum}         
\usepackage{graphicx}
\usepackage[numbers,sort&compress]{natbib}
\usepackage{doi}

\title{AI‑Guided Design and Optimization of Graphite‑Based Anodes via Iterative Experimental Feedback}

\date{}

\usepackage{authblk}

\newbox{\orcid}\sbox{\orcid}{\includegraphics[scale=0.06]{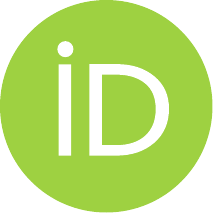}} 
\author[1]{%
	Qian Du\thanks{\texttt{qian.du@hte-company.de}}}%
\author[2]{%
	Mark M. Sullivan}%
\author[2]{%
	James E. Saal}%
\author[1]{%
	Florian Huber}%
\affil[1]{hte GmbH, Kurpfalzring~104, 69123~Heidelberg, Germany}
\affil[2]{Citrine Informatics, 2629~Broadway~St., Redwood~City, CA~94063, USA}

\renewcommand{\headeright}{}
\renewcommand{\undertitle}{}
\renewcommand{\shorttitle}{}

\hypersetup{
pdftitle={A template for the arxiv style},
pdfsubject={q-bio.NC, q-bio.QM},
pdfauthor={David S.~Hippocampus, Elias D.~Striatum},
pdfkeywords={First keyword, Second keyword, More},
}

\begin{document}
\maketitle

\begin{abstract}
This study presents an iterative AI‑guided workflow that accelerates graphite‑based anode development by improving both formulation feasibility and process robustness. Sequential learning via AI/ML-guided multiobjective inverse design for anode optimization was implemented using the Citrine Platform. Starting from a noisy, incomplete dataset, the Citrine Platform was used to generate early surrogate models, which despite low predictive certainty highlighted missing process constraints. By iteratively adding feasibility labels and boundary condition failures, the workflow rapidly converged toward manufacturable, higher‑performing formulations. Fabrication reliability improved from frequent process failures to 100\% successful cell production, while the fraction of cells delivering $\geq$350~mAh~g$^{-1}$ increased from 28.4\% to 84.8\%, with capacity retention rising from 42.1\% to 97.3\%. These results demonstrate that structured, feedback‑driven AI workflows can transform imperfect industrial data into actionable guidance, enabling faster, more reproducible optimization of battery electrode manufacturing.  
\end{abstract}


\section{Introduction}
Battery technology is a cornerstone of modern energy systems, underpinning the deployment of electric vehicles, the integration of renewable energy sources, and the widespread use of portable electronic devices. Among the various battery technologies, lithium-ion batteries (LIBs) have emerged as the dominant solution due to their high energy density, favorable efficiency, and advanced level of commercial maturity. \citep{McIlwaine2021} As global energy demand continues to rise and decarbonization targets become increasingly stringent, further improvements in battery performance, safety, and sustainability remain critical. \citep{Yang2025} Addressing these challenges requires sustained innovation as well as a deeper, system-level understanding of battery materials, formulations, and manufacturing processes. \citep{Drakopoulos2021} Consequently, accelerating battery research and development is no longer optional, but a prerequisite for both technological progress and environmental impact mitigation. \citep{McIlwaine2021}

Conventional battery research workflows typically rely on sequential experimentation, in which individual variables are varied one at a time while all others are held constant. \citep{Drakopoulos2021,Han2025,Duquesnoy2020} Although this approach enables controlled hypothesis testing, it is inherently limited when applied to complex, multi-parameter systems such as battery electrodes. Interactions between composition, processing, and microstructure are often nonlinear and coupled, making it difficult to identify dominant performance drivers or to extrapolate findings beyond narrow experimental windows. As a result, progress under such paradigms is frequently incremental. \citep{Drakopoulos2021,Han2025}

To overcome these limitations, two methodological advances have gained increasing attention in battery research: high-throughput experimentation (HTE) and artificial intelligence / machine learning (AI/ML). \citep{Ismail2025, Nwabara2025, Wang2025, Ng2023} HTE enables the parallelized and reproducible generation of experimental data at scale, significantly increasing experimental throughput while reducing material and time consumption per data point. In parallel, AI/ML within materials informatics platforms provide the capability to extract process--structure--property relationships from large and complex datasets, to explore high-dimensional design spaces, and to guide experimental decisions in an adaptive, data-driven manner. \citep{Han2025,Nwabara2025,Wang2025} By representing complex physical systems with tractable surrogate models, this method enables systematic prediction and optimization of key performance metrics with improved efficiency and scalability. When combined, HTE and AI/ML facilitate a fundamental shift in battery research workflows—from linear, manually guided experimentation toward closed-loop, feedback-rich, and increasingly autonomous research strategies. \citep{Benayad2022}

Among these methods, sequential learning (SL), also referred to as active learning, provides a particularly effective strategy for iterative and data-efficient optimization in high-dimensional design space. In this study, we utilized sequential learning based on the ``forests with uncertainty estimates for learning sequentially'' (FUELS) framework,\citep{ling2017} as implemented in the Citrine Platform. The FUELS framework is based on ensembles of random forest decision trees\citep{HO1998,Breiman2001} and incorporates uncertainty estimation via the infinitesimal jackknife approach,\citep{Wager2014} enabling informed decision-making under limited and noisy data conditions. This framework has been shown to identify top-performing materials with comparable performance and higher computational efficiency than Bayesian optimization methods, and significantly better performance than random-search benchmarks across a wide variety of materials systems.\citep{ling2017,oliynyk2016,fong2021,antono2020,meredig2018,sparks2016,hegde2024} The SL process is iterative and consists of the following steps: (1) training a machine learning model to predict relevant properties with associated uncertainties; (2) identifying promising candidates using a suitable selection strategy (acquisition function); (3) evaluating selected candidates through experiments and/or simulations; and (4) incorporating the results into the dataset and retraining the model. These steps are repeated until one or more high-performing materials are identified or the optimization objectives are achieved.

Within this context, graphite-based anode formulation represesnts a particularly challenging and high-dimensional design space. Electrochemical performance is governed by coupled interactions between active materials, inactive components, and electrode preparation parameters, giving rise to numerous trade-offs that are combinatorially intractable through conventional experimental approaches.\citep{Wang2025,Benayad2022,Zhao2025} This complexity makes anode recipe optimization well suited for AI/ML-guided methodologies.  In this study, we applied the Citrine Platform to guide graphite-based anode formulation, leveraging its ability to identify latent patterns and generate data-driven experimental recommendations. The objective of this work is to share practical insights into the implementation of AI-enabled workflows in battery materials research, using anode formulation optimization as a representative industrially relevant case. While the present study focuses on AI-guided experimentation, future work will explore the tighter integration of AI with high-throughput experimental platforms to further accelerate innovation.

\section{Design of Experiment}
The objective of this study was to systematically explore the formulation space of graphite-based anodes using AI-guided experimental design. By considering compositional variables and processing parameters, the study aims to elucidate how formulation and manufacturing choices influence electrode-level electrochemical performance, and to identify opportunities for performance improvement through data-driven optimization. 

The defined formulation space encompassed a broad and deliberately selected set of variables to ensure sufficient data richness for AI/ML modeling. These variables as depicted in Figure 1 included binder type, solvent type, solid content, and a range of processing parameters such as mixer type, mixing power, mixing time, calendaring pressure, and additional preparation conditions. Together, these parameters were selected to capture formulation–process interactions and their direct impact on key electrode performance metrics. 

\begin{figure}[htbp]
  \centering
  \includegraphics[width=\textwidth]{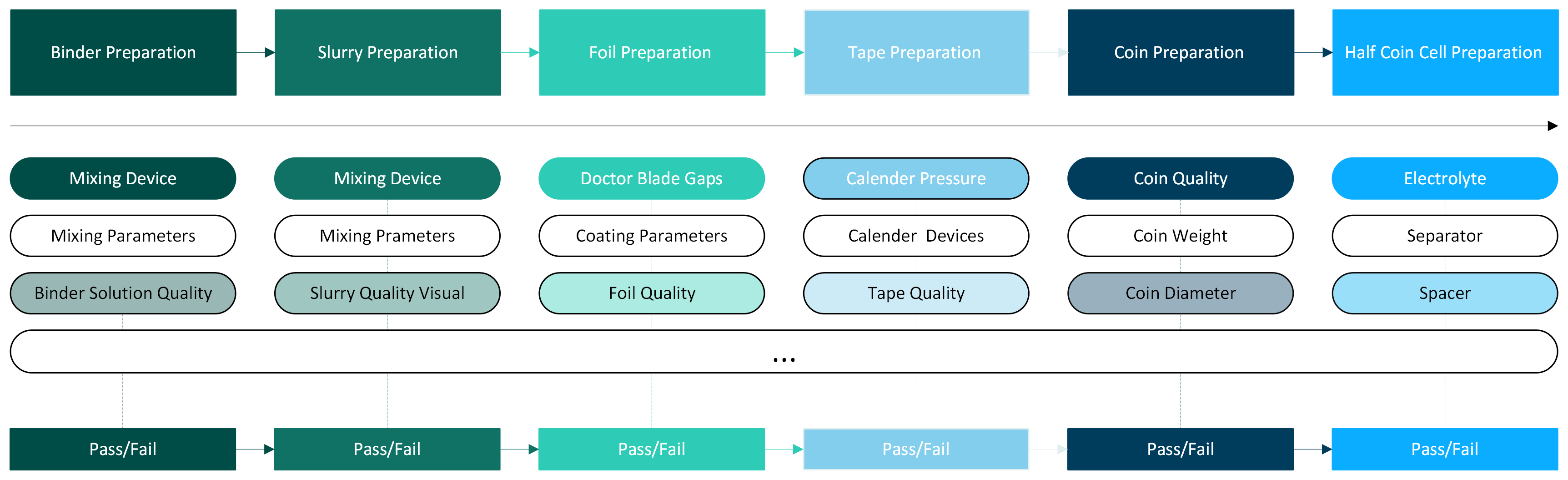}
  \caption{Overview of processing parameters across cell preparation steps.}
  \label{fig:processing}
\end{figure}

In the initial phase of the study, a baseline dataset was assembled from a prior investigation focused on graphite anode systems. The experimental detail is mentioned in section 5. This dataset comprised 12 distinct slurry formulations, each representing a unique combination of compositional and processing parameters. As illustrated in Figure 2, each coin half-cell fabricated in this study can be traced back through a well-defined material history, spanning raw materials, binder preparation, slurry formulation, electrode processing, and final cell assembly. From these slurries, a total of 96 coin half cells, including repeated cells for selected formulations, were fabricated and subjected to electrochemical testing. The resulting dataset served as the foundation for subsequent AI-driven analysis.

\begin{figure}[htbp]
  \centering
  \includegraphics[width=\textwidth]{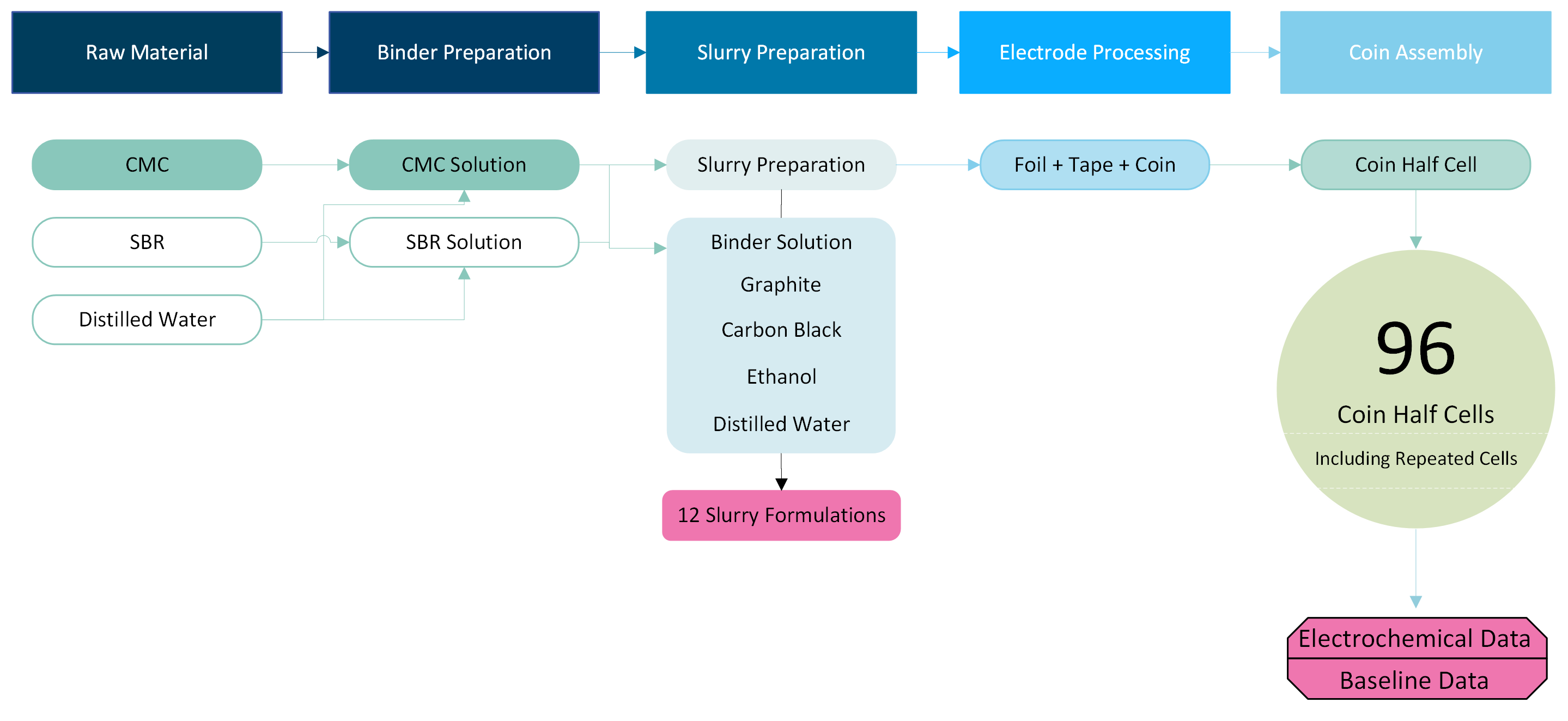}
  \caption{Material lineage of a coin half/cell from raw materials to electrochemical testing and the overview of the electrochemical performance of the raw dataset.}
  \label{fig:processing}
\end{figure}

For each data point, comprehensive end-to-end metadata from raw material specifications through electrode fabrication and cell assembly were systematically recorded and managed using the HTE Laboratory Information Management System (LIMS). This ensured secure data storage, full traceability, and consistent documentation via the myHTE lab software environment. To enable fair and meaningful performance comparisons, all coin cells were evaluated using a standardized electrochemical testing protocol consisting of two formation cycles at 0.1C, followed by two cycles at 0.2C, and 26 subsequent cycles at 0.3333C. Key performance indicators (KPIs), including cycle life, initial discharge capacity, coulombic efficiency at the 5th cycle, and discharge capacity at the 30th cycle, were extracted to construct a consistent baseline dataset for AI model training. 

Up to this stage, all experimental activities and data handling were performed independently of the Citrine Platform. The curated dataset was subsequently uploaded to Citrine Platform, where a complete AI-guided workflow as shown in Figure 3 was executed. This workflow involved formatting the dataset according to a standardized Excel template provided by Citrine, followed by ingestion into the Citrine Platform to generate the initial training set. Based on this training set, a customized AI/ML model was constructed, search spaces were defined by expert scientists to capture the high dimensional parameter range of interest over which inverse-design-guided optimization workflows would explore, and the performance of candidate experiments could be ranked in comparison to predicted performance in relation to user-defined objectives and constraints while incorporating quantified uncertainties. 

\begin{figure}[htbp]
  \centering
  \includegraphics[width=\textwidth]{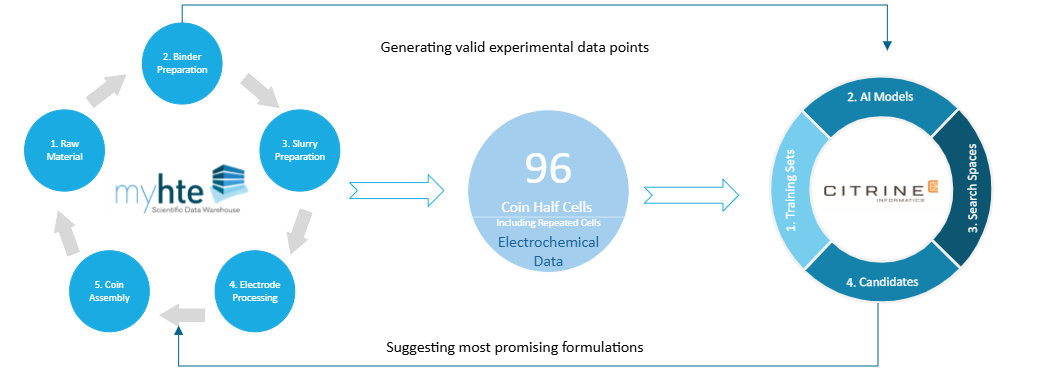}
  \caption{Overview of the AI-guided experimental workflow applied in this study, showing the iterative connection between experimental data, model development, candidate generation, and experimental validation.}
  \label{fig:processing}
\end{figure}

\section{Iterative AI-Guided Experimentation}
This section presents the iterative AI-guided experimentation workflow applied to graphite-based anode formulation optimization. Rather than highlighting a single successful optimization outcome, the emphasis is placed on the process of sequential learning—including failed experiments, data restructuring, constraint refinement, and model–experiment co-evolution. This reflects the reality of early-stage materials development, where learning efficiency and knowledge codification are as critical as absolute performance gains.

\subsection*{\normalfont\bfseries Iteration 0: Baseline Modeling and Diagnostic Assessment}
The initial interaction with the AI platform began by processing the baseline dataset through a complete modeling cycle as shown in Figure 3 consisting of: (i) training dataset preparation, (ii) AI model training, (iii) definition of the formulation and process search space, and (iv) candidate generation and ranking (the entire four step process took ca.\ 30--60 min).  

The baseline dataset, derived from prior binder-focused studies, exhibited substantial variability across all electrochemical key performance indicators, including discharge capacity, coulombic efficiency, and cycle stability. As illustrated in Figure 4, both absolute performance and cell-to-cell reproducibility were poor. Importantly, this dataset was intentionally retained without manual curation or exclusion of low-quality data points. The objective at this stage was not to maximize predictive accuracy, but to evaluate how the AI framework interprets and responds to a realistic, noisy experimental starting point—a scenario frequently encountered in early formulation development or technology transfer contexts.

\begin{figure}[htbp]
  \centering
  \includegraphics[width=\textwidth]{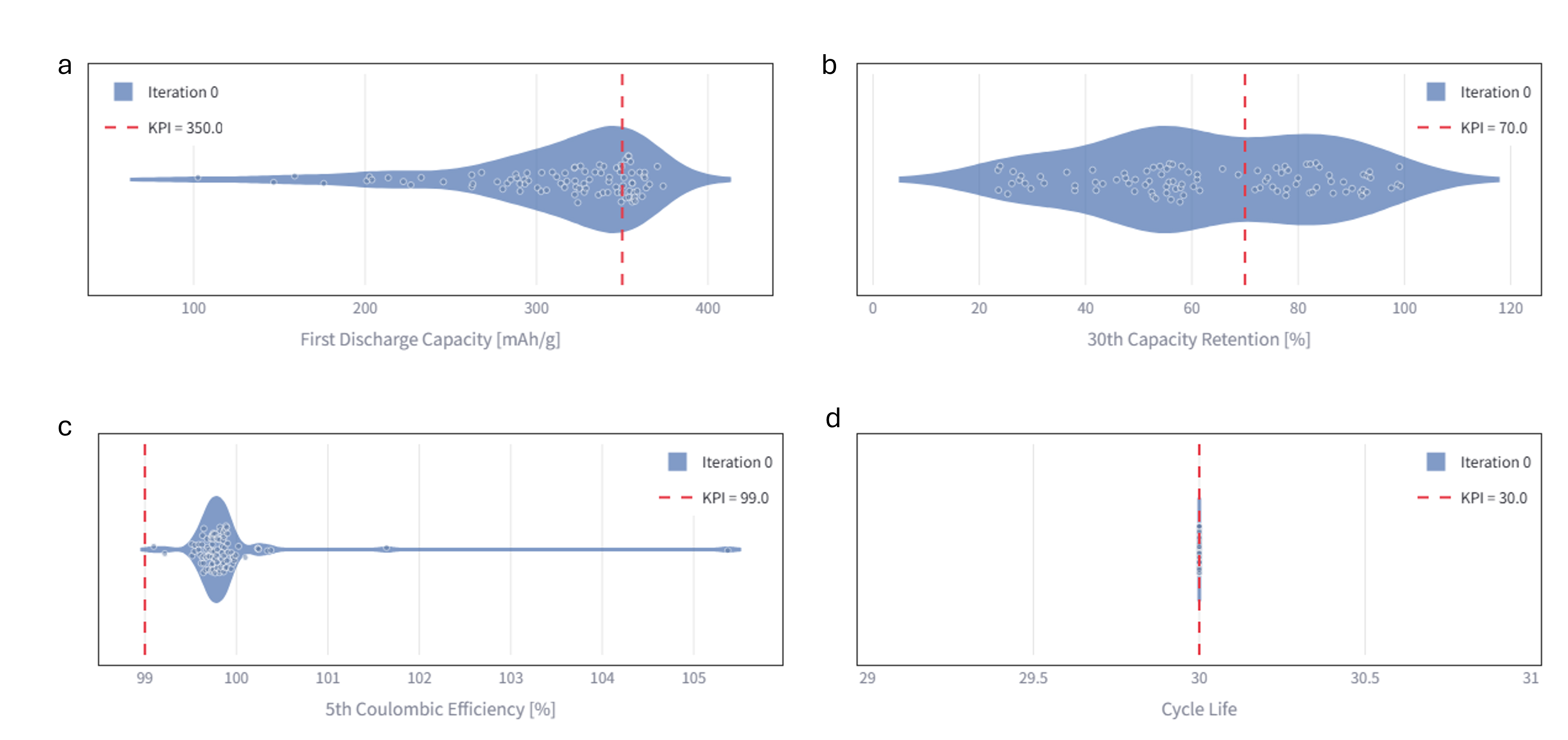}
  \caption{Distribution plots of electrochemical key performance indicators for the baseline coin half-cell dataset, including, first discharge capacity, fifth-cycle coulombic efficiency, capacity retention after 30 cycles and cycle life. Dashed vertical reference red lines indicate predefined key performance indicators (KPIs) used as decision thresholds during candidate selection.}
  \label{fig:processing}
\end{figure}

Despite the limited data quality, the AI model was able to infer preliminary relationships between formulation variables (e.g., binder type, solvent composition, and solid content) and electrochemical performance, as depicted in Figure 5. Resulting candidate formulations exhibited likelihood-of-improvement (LoI) scores\citep{ling2017} on the order of 0.16. Likelihood of improvement is a representative metric that can be thought of as a probability of achieving all target objectives and constraints. Any positive value for a likelihood of improvement implies feasibility of improving materials and model prediction in the direction of interest, with higher scores approaching to 1 representing a higher likelihood of achieving all objectives. This single metric lumps together factors related to all objectives, constraints, the associated uncertainties of the predictions of those targets, and the user-specified baselines of those performance metrics. 

This result suggests that given the high uncertainty associated with the current dataset and the broadly defined search space, the probability of identifying formulations that deliver substantial improvements while simultaneously satisfying all four KPIs is low at this stage.

\begin{figure}[htbp]
  \centering
  \includegraphics[width=\textwidth]{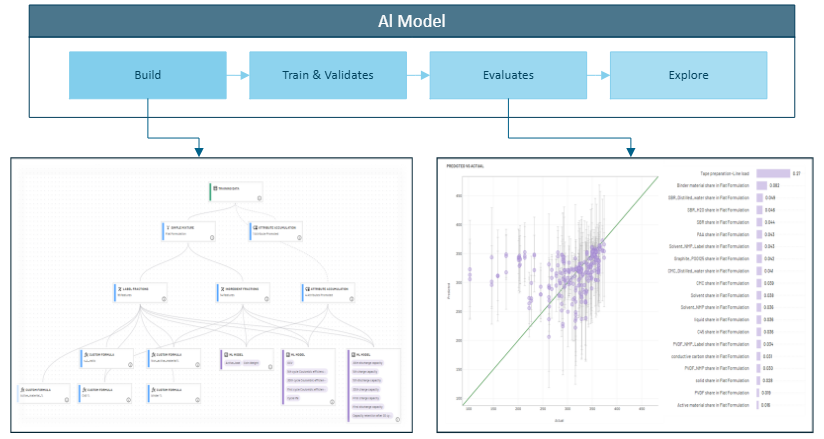}
  \caption{Schematic of AI model construction and evaluation within the Citrine Platform. The upper panel outlines the model lifecycle implemented in this study, including model building, training and validation, evaluation, and candidate exploration. The lower left panel presents a representative view of the constructed AI model linking formulation ingredients and processing parameters to electrochemical performance targets. The lower right panel shows the model evaluation outputs, including a predicted-versus-experimental comparison and a feature-importance ranking that summarizes the relative contribution of individual input variables to the model predictions. All visualizations shown are generated directly by the Citrine Platform based on the specified model configuration.}
  \label{fig:processing}
\end{figure}

At the initial modeling stage (Iteration~0), the workflow outputs were interpreted primarily as diagnostic indicators rather than reliable predictions. High uncertainty reflects gaps in data coverage, lack of domain knowledge incorporation into model architecture and feature engineering, and limitations in how the formulation space and constraints had been defined. The first iteration with the preliminary model therefore helped identify where additional descriptors, domain knowledge, or refined boundaries were needed to better represent the graphite anode system. This stage also reflects a common reality in early-stage research and development projects, where work often begins with nonideal or incomplete datasets. The purpose of this stage was not to deliver optimal formulations but to establish a structured baseline from which experimental learning could progress. Therefore, rather than treating model accuracy as the ultimate measure of success, the key objective was to use both predictions and quantified uncertainty to determine where new experiments would provide the most informative data.

\subsection*{\normalfont\bfseries Iteration 1: First AI-Suggested Experiments and Failure-Driven Learning}

Based on the initial AI recommendations, five candidate formulations were manually selected for experimental validation (Table 1). This selection involved expert intervention encoded as domain knowledge into Search Spaces in the platform to exclude implausible recipes, such as binder solution concentrations that are not achievable in practice (e.g., excessively high CMC concentrations that cannot form homogeneous slurries). Selected candidates were synthesized and evaluated experimentally with additional calendaring pressures applied to increase data density.

\begin{table}[htbp]
\centering
\caption{Composition of AI-suggested candidate formulations evaluated in Iteration~1.}
\label{tab:iteration1}

\begin{tabular}{lccccccc}
\toprule
 & \multicolumn{7}{c}{\textbf{Formulation in myhte}} \\
\cmidrule(lr){2-8}
 & C45 & Graphite\_PO0125 & SBR & CMC & Distilled\_water & Ethanol & Solid ratio \\
\midrule
V1\_Candidate 1 & 1.22 & 98.30 & 0.00 & 0.48 & 100.00 & 0.00 & 0.57 \\
V1\_Candidate 2 & 1.36 & 95.15 & 3.49 & 0.00 & 100.00 & 0.00 & 0.50 \\
V1\_Candidate 3 & 11.68 & 83.81 & 4.49 & 0.00 & 0.00 & 100.00 & 0.61 \\
V1\_Candidate 4 & 0.80 & 95.15 & 3.97 & 0.08 & 50.02 & 49.98 & 0.40 \\
V1\_Candidate 5 & 0.60 & 95.38 & 4.02 & 0.04 & 50.10 & 49.90 & 0.40 \\
\bottomrule
\end{tabular}
\end{table}

As shown in Figure 6, four of the five candidate materials failed at various stages along the manufacturing process, resulting in little usable data for those materials. The unusually high failure rate observed in this iteration suggests that the baseline data used for prediction was not sufficiently representative of the practical fabrication constraints. Figure 6 makes these failures explicit, spanning multiple stages of the manufacturing workflow, and this prompted a closer examination of what information was missing from the initial dataset. Such outcomes commonly reflect either (i) incomplete incorporation of expert domain knowledge into the formulation of the search‑space constraints, and/or (ii) insufficient prior exploration of the explicit feasibility limits of real manufacturing process parameters. Notably, in this case none of the baseline records explicitly encoded failure-related outcomes; as a result, the model was trained almost exclusively on “successful” trajectories and lacked the information needed to learn feasibility boundaries. In this context, failure should not be treated as an experimental dead-end to be discarded. Instead, it serves as a constructive signal—an important boundary condition that distinguishes manufacturable from non-manufacturable formulations and should be explicitly captured and incorporated into subsequent modeling cycles. 

\begin{figure}[htbp]
  \centering
  \includegraphics[width=0.9\textwidth]{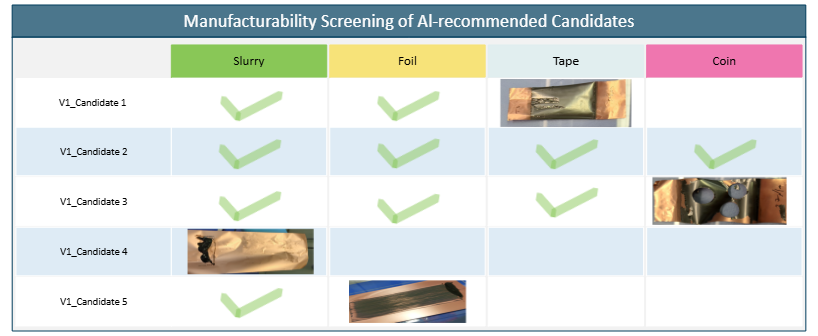}
  \caption{Experimental outcomes of Al-suggested candidates formulations across the fabrication workflow in Iteration 1.}
  \label{fig:processing}
\end{figure}

\subsection*{\normalfont\bfseries Iteration 2: Data Structure Augmentation and Constraint Refinement}

In response to the insights gained from Iteration 1, the dataset and modeling framework were substantially revised. New process-level labels were introduced to explicitly capture experimental feasibility and quality, including indicators for slurry homogeneity, coating success, calendaring integrity, and overall fabrication pass/fail status (Figure 1). These labels are rarely reported in conventional battery research publications, yet they encode critical practical knowledge that governs experimental success. 

In parallel, additional boundary-condition data points were deliberately added, representing known infeasible binder and solvent combinations. This augmentation was designed to improve the model’s ability to discriminate between feasible and infeasible regions of the formulation space, rather than merely interpolating performance trends within a narrow success domain. 

Following dataset restructuring, the AI workflow (Figure 3) was executed once more with refined search space constraints, including explicit limits on binder solution concentrations and solid-to-liquid ratios. The KPI set was simplified from six metrics to four core indicators to reduce objective complexity at this stage of learning as shown in Figure 4. 

As a result, the newly generated candidates in this iteration 2 exhibited new scores, reaching values up to approximately 0.33. Although predictive uncertainty remained high—reflecting the still-limited dataset size—the candidate formulations were markedly more realistic from a processing standpoint, and five candidates as can be seen in Table 2 were selected for experimental validation. Notably, LoI scores are not directly comparable across different candidate generations when the underlying objectives and constraints differ and as statistical ML models are updated with additional training data. In subsequent experimental iterations, where the key performance indicators and constraint definitions are held fixed, LoI values can be meaningfully compared and used to assess relative improvements across candidate formulations. 

\begin{table}[htbp]
\centering
\caption{Composition of AI-suggested candidate formulations evaluated in Iteration~2.}
\label{tab:iteration2}

\begin{tabular}{lccccccc}
\toprule
 & \multicolumn{7}{c}{\textbf{Formulation in myhte}} \\
\cmidrule(lr){2-8}
 & C45 & Graphite\_PO0125 & SBR & CMC & Distilled\_water & Ethanol & solid ratio \\
\midrule
V2\_Candidate 1 & 1.164 & 93.76  & 3.039 & 2.037 & 80.54 & 19.46  & 0.30 \\
V2\_Candidate 2 & 1.578 & 93.367 & 3.026 & 2.028 & 88.47 & 11.53  & 0.30 \\
V2\_Candidate 3 & 4.561 & 91.21  & 2.578 & 1.646 & 100.0 & 0.00   & 0.50 \\
V2\_Candidate 4 & 1.024 & 91.88  & 4.200 & 2.890 & 100.0 & 0.00   & 0.30 \\
V2\_Candidate 5 & 7.188 & 86.68  & 3.594 & 2.537 & 68.92 & 31.079 & 0.46 \\
\bottomrule
\end{tabular}
\end{table}

In the second experimental iteration, all AI-selected formulations were successfully fabricated into coin cells without process failure. This alone represented a significant milestone, demonstrating that the incorporation of expert domain knowledge in the form of feasibility labels and refined constraints effectively aligned model recommendations with laboratory reality.

Electrochemical testing indicated clear improvements in both performance and reproducibility relative to the baseline dataset and earlier iterations, as shown in Figure~7. After two rounds of the AL/ML feedback loop, a substantially larger fraction of candidates met the defined KPIs: 84.8\% of cells achieved a first discharge capacity of 350~mAh\,g$^{-1}$, compared with only 28.4\% in the baseline dataset, as shown in Figure~7a. Capacity retention performance increased similarly; initially, 42.1\% of baseline materials exceeded the baseline KPI performance, whereas 97.3\% of Iteration~2 candidates exceeded the KPI baseline, as shown in Figure~7b. By contrast, cycle life and fifth-cycle coulombic efficiency contributed less diagnostic value at this stage, as their baseline values were already high and exhibited limited variation. To more effectively evaluate coulombic efficiency in future iterations, implementing explicit upper and lower constraints would provide clearer discrimination and enhance their utility within the optimization framework.

\begin{figure}[htbp]
  \centering
  \includegraphics[width=\textwidth]{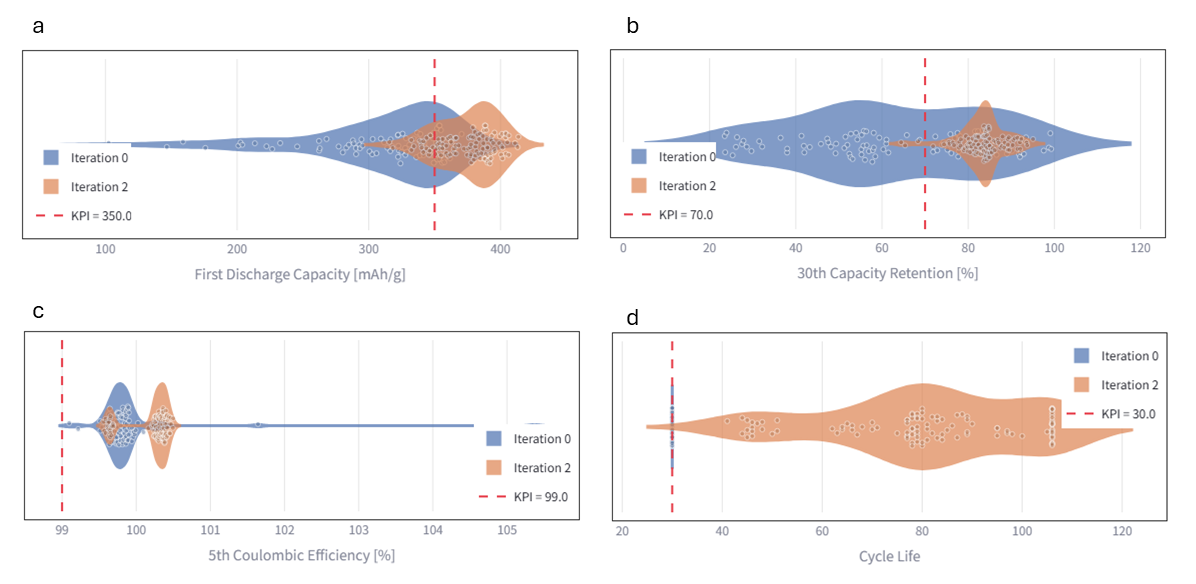}
  \caption{Distribution plots of electrochemical key performance indicators for the baseline coin half-cell dataset and the result from iteration 2, including, first discharge capacity, fifth-cycle coulombic efficiency, capacity retention after 30 cycles and cycle life. Dashed vertical reference red lines indicate predefined key performance indicators (KPIs) used as decision thresholds during candidate selection. Blue color represents iteration 0, the basline dataset. Orange color represents the result from iteration 2.}
  \label{fig:processing}
\end{figure}

Figure~8 presents the electrochemical results of the V2\_candidates 5 evaluated under different calendaring pressures. As shown in Figure 8, averaging the electrochemical performance of replicate cells fabricated on the same tape enabled a clearer assessment of the influence of calendaring pressure. This observation highlights the importance of generating high--quality and reproducible datasets, which are essential for reliably analyzing the effects of individual process parameters and for guiding subsequent research directions. Notably, the influence of calendaring pressure--previously inferred implicitly by the AI model through earlier iterations--became experimentally observable in this iteration. This consistency between model--guided insights and experimental outcomes provides supporting evidence for the credibility of the learned process--structure--property relationships.

\begin{figure}[htbp]
  \centering
  \includegraphics[width=\textwidth]{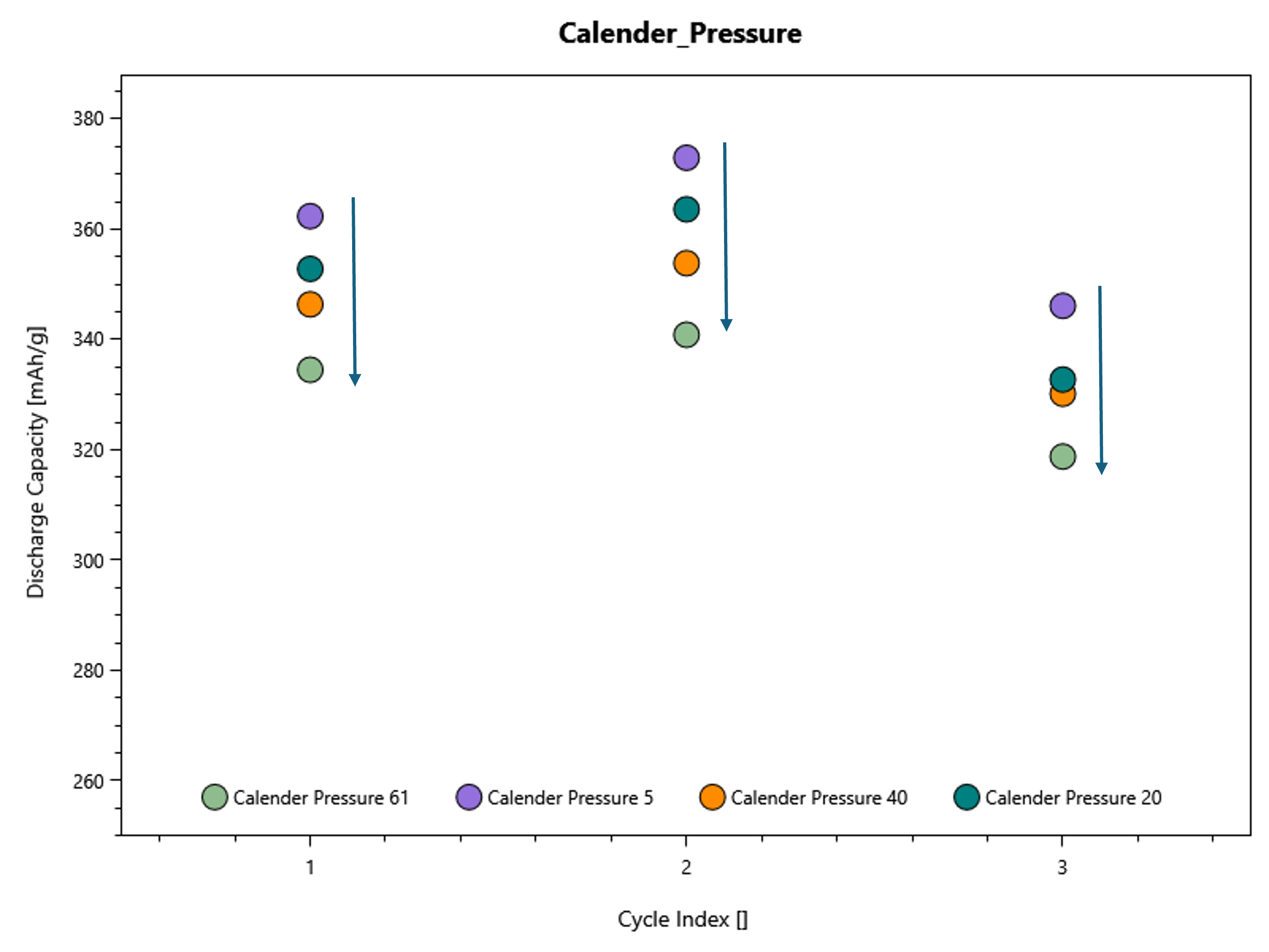}
  \caption{Electrochemical performance of AI-selected Candidate~4 tapes in experimental Iteration~2. Plotted is the average discharge capacity versus cycle number for a representative tape processed under different calendaring pressures 5,20, 40, 61 N/mm. Each data point reflects the mean performance of multiple cells fabricated from the same tape under the same pressure condition, highlighting the systematic dependence of electrochemical behavior on calendaring pressure.}
  \label{fig:figure8}
\end{figure}

\subsection*{\normalfont\bfseries Iteration 3: Convergence Toward Feasible and Reproducible Formulations }

After incorporating the newly generated experimental data from iteration 2, the AI-guided workflow was re-executed, resulting in a marked increase in the likelihood-of-improvement (LoI) scores to ~0.83 in comparison to LoI scores on the order of 0.33 for the previous iteration which had identical performance objectives and constraints, as depicted in Figure 9. Crucially, this improvement was not limited to numerical optimization metrics, but was also realized via experimentally reproducible electrochemical behavior and well-defined process-dependent trends. Compared to earlier iterations (Figure 10), the candidates generated in Iteration 3 were able to satisfy all defined KPIs despite the presence of residual uncertainty, providing a clear and robust signal of optimization success. As a result, candidate selection became significantly more straightforward and better supported by quantitative evidence. At this stage, the model outputs were underpinned by a substantially stronger experimental foundation gathered by efficient AI/ML guided candidate sample selection, enabling increased confidence in the inferred process–performance relationships. 

Overall, this progression illustrates how iterative AI/ML-guided, expert-driven data enrichment can transform intuition-driven or simple statistical experimental design approaches into a more reliable, systematic and informative framework for efficiently improving understanding and optimizing battery manufacturing processes.

\begin{figure}[htbp]
  \centering
  \includegraphics[width=\textwidth]{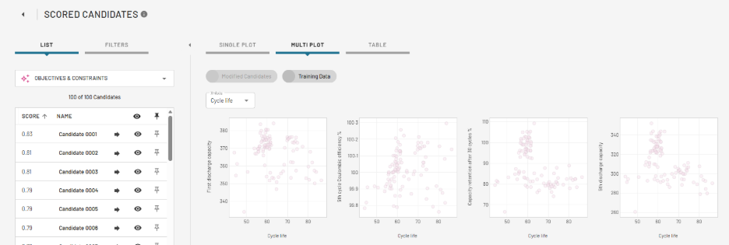}
  \caption{Citrine Al-guided suggestion based on thirded iteration.}
  \label{fig:figure8}
\end{figure}

\begin{figure}[htbp]
  \centering
  \includegraphics[width=\textwidth]{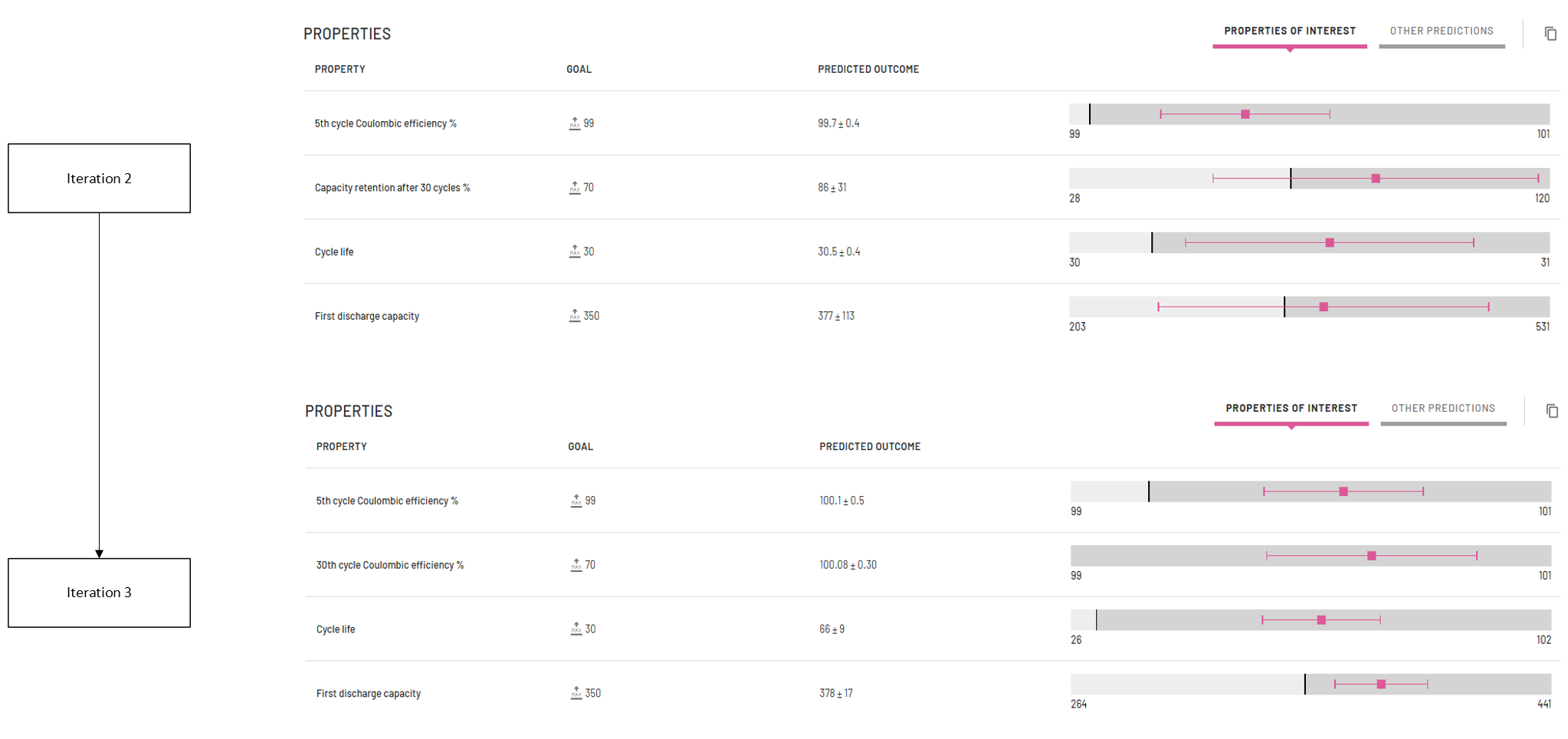}
  \caption{Comparison of AI-identified best candidates across key electrochemical gs in iterations 2 and 3.}
  \label{fig:figure8}
\end{figure}

\section{Material and Methods}

\subsection{Anode Slurry Preparation}

Natural graphite (MSE Supplies) and conductive carbon (C45) were used as the active material and conductive additive, respectively. Sodium carboxymethyl cellulose (Na-CMC, battery grade) and Licity\textsuperscript{\texttrademark}~2698~XF (50~wt\% SBR in water) were employed as optional binder materials. Distilled water and ethanol served as optional solvents. Each slurry formulation contained the active material, conductive carbon, at least one binder, and one solvent. The ratios of all components followed the candidate formulations generated by the Citrine Platform.

Two mixing procedures were used depending on whether SBR was included. For formulations using Na-CMC as the only binder, all components were mixed in a Thinky ARE-250 planetary mixer at 2000~rpm for 20~min. For formulations containing SBR, all ingredients except the SBR component were first mixed at 2000~rpm for 15~min; the required amount of Licity\textsuperscript{\texttrademark}~2698~XF was then added, followed by an additional 5~min of mixing at 500~rpm to ensure proper incorporation of the SBR binder.

\subsection{ Electrode Preparation}

The prepared anode slurry was coated onto copper foil using an Erichsen
Coat Master with a fixed 60~$\mu$m doctor-blade gap and a coating speed
of 5~mm~s$^{-1}$ in the forward direction. The coated foil was dried in a
Binder VDL-53 oven at 70~$^\circ$C for 12~h to remove the solvent.

Following drying, electrodes were calendered at various line pressures
determined by the Citrine Platform. Calendaring was carried out using a
Sumet CA~5/250 calender machine operated at 25~$^\circ$C with a forward
feed speed of 0.5~m~min$^{-1}$. After calendaring, electrodes were
punched into 15~mm discs using a Nogami handheld punch. The discs were
transferred into an Ar-filled glovebox (O$_2$~$<$~0.1~ppm,
H$_2$O~$<$~0.1~ppm) and vacuum-dried at 120~$^\circ$C for 12~h before
cell assembly.

\subsection{ Coin Cell Assembly}
All coin cell assembly steps were performed inside an Ar-filled glovebox.
CR2032 coin cells were assembled using the prepared graphite electrodes
as the working electrode. A lithium metal chip served as the counter
electrode, and a GF/D glass fiber separator was placed between the
electrodes. Each cell was filled with 100~$\mu$L of LP30 electrolyte
(1~M LiPF$_6$ in EC/DMC, 1:1~v/v) to ensure adequate wetting.

Cells were sealed using a Hohsen HSACC crimping machine under controlled
pressure conditions. Assembled cells were rested for 12~h prior to
electrochemical testing to ensure full electrolyte penetration.

\subsection{ Coin Cell Testing}
Electrochemical measurements were performed on an Arbin cycler. Cells
were cycled between 0.005~V and 1.50~V vs.\ Li/Li$^{+}$. The formation
protocol consisted of two cycles at C/10, followed by two cycles at C/5,
and then 26 cycles at C/3. For cells not belonging to the baseline
dataset, cycling was not terminated after the initial 26 cycles at C/3.

\subsection{ Citrine Platform Overview}
The Citrine Platform was used to conduct sequential learning–guided materials optimization through an iterative closed-loop workflow integrating data management, machine learning, design-space definition, and candidate selection. Experimental and computational data were structured using the Graphical Expression of Materials Data (GEMD) framework, which encodes materials provenance, processing history, measurement metadata, and uncertainty in a graph-based representation suitable for machine learning applications. Candidate materials and process conditions were featurized using composition-, process-, and formulation-aware descriptors, including chemically informed features derived from stoichiometry and domain-specific processing variables. Predictive models were constructed using the Citrine Platform’s automated machine learning framework, with random forest ensembles serving as the primary surrogate model due to their robustness for heterogeneous materials datasets and compatibility with uncertainty quantification. Model uncertainties were estimated using jackknife-based variance methods with explicit bias correction, enabling calibrated heteroscedastic uncertainty estimates appropriate for sequential learning. Design spaces were constrained using domain knowledge, including compositional bounds, mixture constraints, optional ingredients, and process-variable limits, ensuring that generated candidates were physically realizable and experimentally relevant. Sequential learning iterations were performed by sampling the constrained design space, predicting target properties and associated uncertainties, and ranking candidates using acquisition functions balancing exploration and exploitation, including Likelihood of Improvement. Selected candidates were experimentally evaluated, and the resulting measurements were iteratively incorporated into the training dataset to refine subsequent model predictions and accelerate convergence toward optimal materials performance.

\section{Conclusions}
This case study demonstrates that AI‑guided experimentation, when combined with structured high‑throughput workflows, can significantly accelerate battery materials development even under realistic constraints such as limited data, complex processing parameters, and frequent experimental failure. Early iterations demonstrated that AI/ML guided optimization can highlight the detrimental impacts of the common habits of neglecting feasibility- and process‑related outcomes within materials development datasets. Recognizing experimental failure as valuable information, rather than noise to be discarded, proved essential. By explicitly capturing failure modes and incorporating feasibility constraints, the workflow identified meaningful formulation boundaries and aligned model recommendations with manufacturing reality. 

Quantitatively, the iterative refinement delivered substantial gains. The likelihood‑of‑improvement (LoI) for candidate formulations, a composite statistical metric representing a pseudo-probability of achieving targets of interest, increased from smaller than 0.33  to 0.83 betwen the second and third experimental iterations—a nearly threefold improvement in performance. Fabrication reliability improved from initial process failures to 100\% successful coin‑cell production for all AI‑suggested candidates. Electrochemical performance also advanced, with Iteration 3 cells exhibiting higher first‑cycle discharge capacities, improved coulombic efficiency, and reproducible calendaring‑pressure trends that were absent in the baseline dataset. Rolling‑update validation confirmed that even under data scarcity, iterative learning systematically narrowed model uncertainty and strengthened predictive reliability. 

Together, these results show that iterative, data-enriched AI workflows do not need perfect accuracy to create value—they need structure, feedback, and continuity. AI-guided experimentation can effectively and efficienctly run the gamut of exploration to exploitation based on the current state of the data and optimization targets, maximizing the efficiency of expert resources devoted to experimentation. Ultimately, this work provides clear evidence that better‑defined search spaces lead to better models, better candidates, and more reproducible, higher‑performing battery materials—establishing a scalable foundation for future AI‑enabled materials development.

\section{Outlook}
The outcomes of this study point toward a highly promising future in which AI‑guided experimentation becomes a core driver of accelerated battery innovation. The iterative gains achieved here – improved feasibility, clearer process–performance relationships, increasingly effective candidate selection, and ultimately improved experimental efficacy – demonstrate that systematically following AI‑generated recommendations can unlock tangible performance improvements, especially when combined with consistent high‑throughput experimentation. 

Looking ahead, expanding this workflow with richer datasets and greater automation will enable AI models to explore broader formulation and process spaces, uncover hidden interactions, and propose candidates that move beyond incremental gains toward genuinely new performance classes. As domain knowledge is further embedded into the models – through refined constraints, nuanced failure modes, and more complete process metadata – AI can begin to reveal mechanistic insights that traditionally require extensive manual experimentation. 

As experimental and modeling loops become more tightly integrated, faster iteration cycles, automated campaign planning, and more autonomous optimization workflows become achievable. Importantly, the methodology demonstrated here is broadly transferable to other electrode chemistries, electrolytes, or entirely different materials systems. 

With continued refinement, the AI–HTE framework established in this work has strong potential to shorten development timelines, reduce uninformative experimentation, and support the discovery of truly best‑in‑class battery materials and technologies.

\bibliographystyle{spmpsci.bst}
\bibliography{references}

\end{document}